\RequirePackage{amsmath}
\documentclass[runningheads,a4paper]{llncs}

\usepackage{amssymb}
\usepackage{graphicx}
\usepackage[caption=false]{subfig}
\usepackage{amssymb}

\usepackage{url}
\urldef{\mailsa}\path|{jordan.henrio, thomas.henn}@cs.osakafu-u.ac.jp|
\urldef{\mailsb}\path|tomoharu.nakashima@kis.osakafu-u.ac.jp|
\urldef{\mailsc}\path|akym@fukuoka-u.ac.jp|   
 
\newcommand{\keywords}[1]{\par\addvspace\baselineskip
\noindent\keywordname\enspace\ignorespaces#1}

\newcommand{\etal}{\textit{et al.}}

\makeatletter
 \def\@textbottom{\vskip \z@ \@plus 20pt}
 \let\@texttop\relax
\makeatother

\begin{document}

\mainmatter  % start of an individual contribution

\title{Selecting the Best Player Formation\\ for Corner-Kick Situations \\
Based on Bayes' Estimation}

\titlerunning{Selecting the Best Player Formation for Corner-Kick Situations
Based on Bayes' Estimation}

\author{Jordan Henrio$^1$\and Thomas Henn$^1$\and Tomoharu Nakashima$^1$\and \\ 
Hidehisa Akiyama$^2$}
\authorrunning{Henrio, Henn, Nakashima and Akiyama}

\institute{$^1$Osaka Prefecture University, Osaka, Japan\\
\mailsa\\
\mailsb\\
$^2$Fukuoka University, Fukuoka, Japan\\
\mailsc\\
}

%\toctitle{Lecture Notes in Computer Science}
%\tocauthor{Authors' Instructions}
\maketitle

\begin{abstract}
In the domain of RoboCup 2D soccer simulation league, appropriate player 
positioning against a given opponent team is an important factor of soccer team 
performance. This work proposes a model which decides the strategy that should 
be applied regarding a particular opponent team. This task can be realized by 
applying preliminary a learning phase where the model determines the most 
effective strategies against clusters of opponent teams. The model determines 
the best strategies by using sequential Bayes' estimators. As a first trial of 
the system, the proposed model is used to determine the association of player 
formations against opponent teams in the particular situation of corner-kick. 
The implemented model shows satisfying abilities to compare player formations 
that are similar to each other in terms of performance and determines the right 
ranking even by running a decent number of simulation games. 
\end{abstract}

\section{Introduction}

One of the essential parts in developing a team in the RoboCup 2D soccer 
simulation league is to design an effective strategy or method that outperforms 
opponent teams. Player formation is one of the most important aspects in the 
strategy design as it gives the guidelines of the decision making during the 
game. The player formations are generally designed according to a given opponent
team. However, this tasks is labourious since the search space can be really 
large depending on the set-play the formation is associated with. In addition, 
selecting the best strategy regarding unknown opponents is one of the most 
challenging task of this league.

On the other hand, it is not necessary to create a specialized player 
distribution against each of all opponents as it is possible that some of them 
are similar regarding particular features. By using this fact, it is possible to 
cluster similar opponents together and then look for the most effective 
strategy against this group.

This research proposes a model which groups similar opponent teams together 
during a learning stage and determines the most effective player 
formation for each cluster by using sequential Bayes' estimations. Then, 
during a real game the system classify the current opponent among one of the 
determined clusters and apply the strategy that has been estimated to be the 
best regarding the resulting classification.

\section{Related Work}

The task of recognizing the opponent strategy in order to apply an 
appropriate counter action has been already adressed in previous researches. 
For example, the works of Visser \etal\cite{annclassification} and
Dr{\"u}cker \etal\cite{virtualwerder} propose a system for recognizing 
opponent's formations and then apply a counter formation. This is done by using 
an artificial neural network that is able to classify data among 16 formation 
classes and then apply the counter formation especially designed against each 
class. Classified data are a representation of the field as a grid expressing 
the formation of the opponent. Riley and Veloso \cite{gridanalysis} also 
proposed a method performing opponent classification by using a grid 
representation of the field. However, the grid is used for displacement and 
location of objects instead of just observing the structures of formations. 
Also, they used a decision tree instead of neural networks. 

However, this paper focuses on how to select the best player formation
regarding a particular cluster rather than the issue of how build clusters. The 
selection issue is a well know problem in probability, often named as the 
$k$-armed bandit problem. This problem has been already addressed in the context 
of simulated soccer game by Bowling \etal \cite{behaviorselection}. They 
proposed to apply an algorithm that selects the most effective and available 
team plans during particular situations. The most effective plan is the one that 
minimizes the regret which is the amount of additional reward that could have 
been received by behaving optimally. The work presented in this paper is similar
in the sense that we also focus on a method which selects the best choice in a 
particular situation. However, the situation is considered to be a particular 
opponent team in a particular event of the game and not a particular state of 
the environment. Doing so allows us to focus on more precise effectiveness 
measurement functions.

\section{Proposed Model}
\begin{figure}[h]
  \centering
  \includegraphics[scale=0.2]{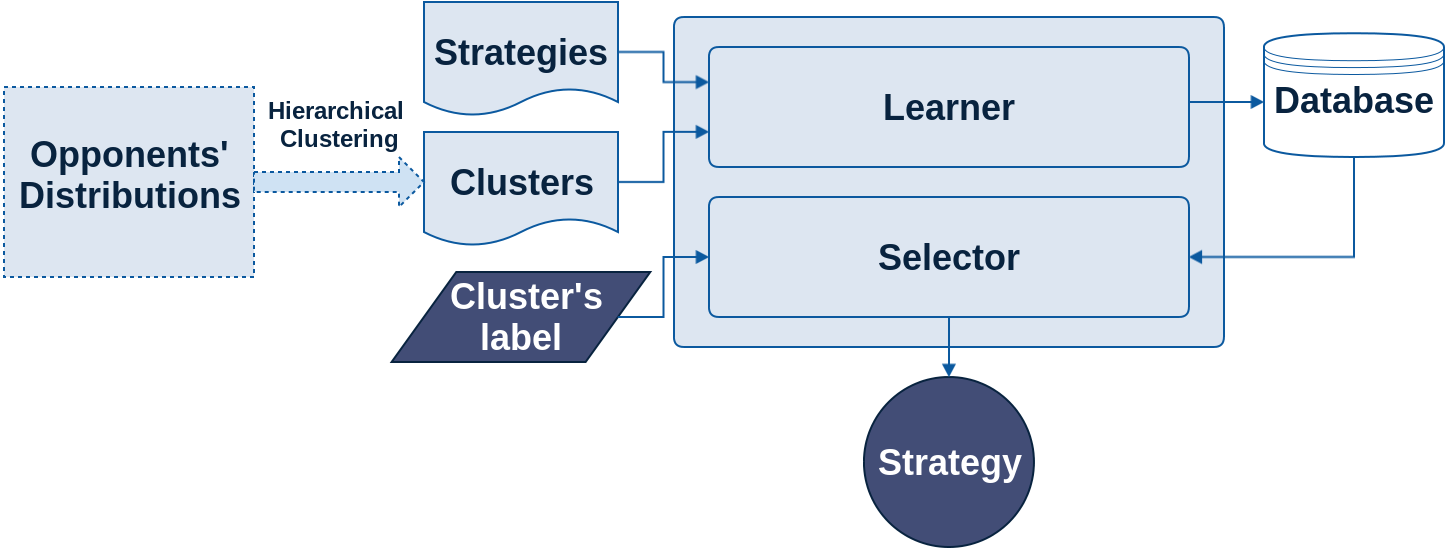}
  \caption{Proposed model.}
  \label{fig:model}
\end{figure}

As a solution, we propose to use a simple model as shown in \figurename 
\ref{fig:model}. This system takes a label of cluster of opponent teams' as an 
input parameter and returns the best strategy to apply. However, it is difficult
to estimate the best strategy among the ones we have in hand. For this 
reason the proposed model consists of two modules, \textit{Learner} and 
\textit{Selector}.

The \textit{Learner} part works in offline mode. It takes a set of clusters as 
an input parameter. Clusters are obtained by applying hierarchical clustering on
opponent teams' distributions before the \textit{Learner} works. Then, its role 
is to learn against each cluster, by looking at the set of strategies that we 
have already developed, which one is the most appropriate. This decision is done
by performing statistical analysis on simulated games by using the different 
strategies as it is explained in Section \ref{sec:learning}. Once the learner is 
able to decide which strategy we should apply regarding a particular cluster of 
opponent teams, it inserts the cluster-strategy pair in a database.

The \textit{Selector} part works in online mode. It takes the resulting 
classification of the current opponent team as input. Then, by using the 
estimations done by the learner it can directly return the best strategy to 
apply.

As a first trial of this system, the proposed model was used to determine which 
corner-kick formations should be used against particular clusters of opponent 
teams from JapanOpen competitions. This championship is the RoboCup yearly 
meeting within Japan.

\section{Opponents Clustering}

\subsection{Team distributions}

\begin{figure}[h]
  \centering
  \includegraphics[scale=0.15]{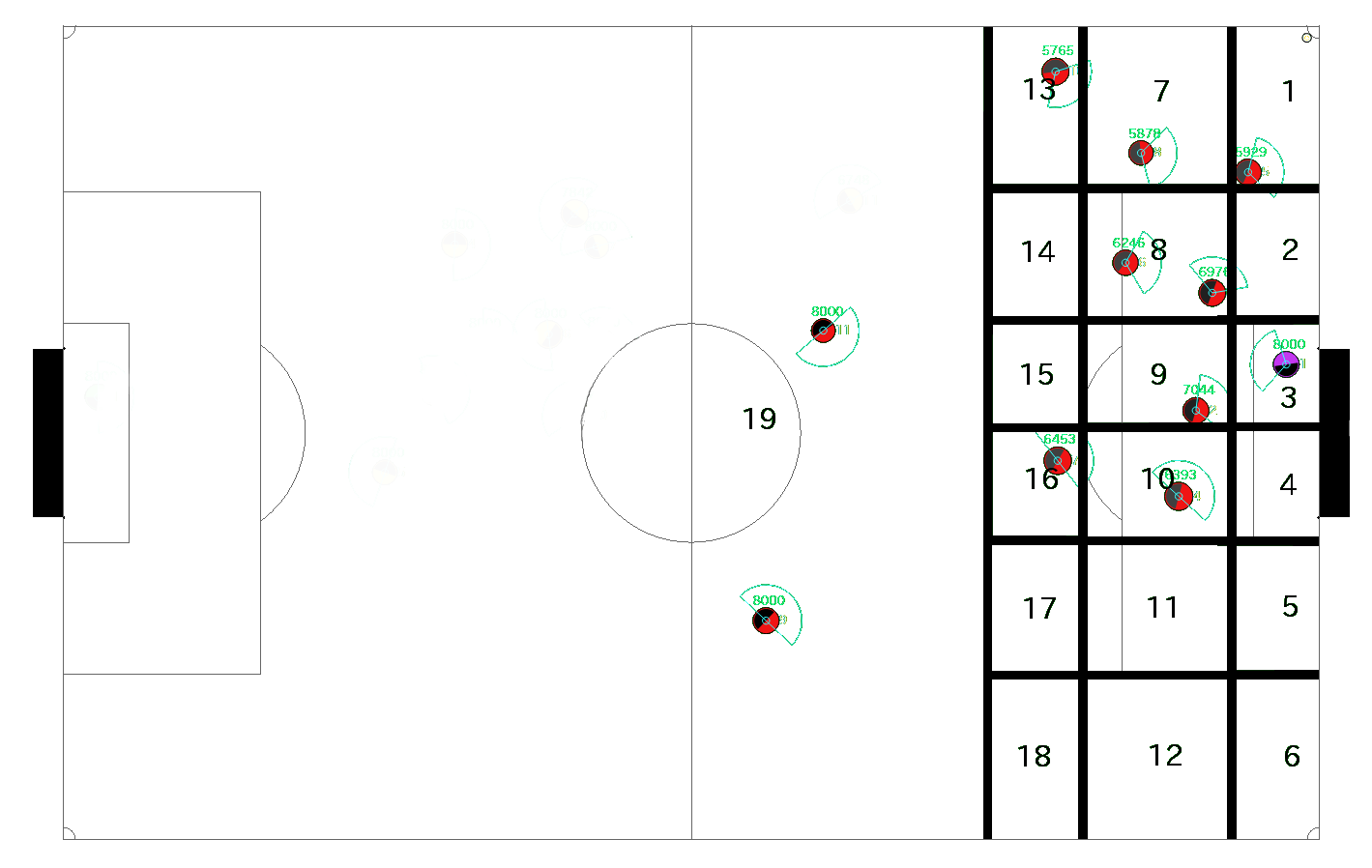}
  \caption{19 blocks of the partitioned soccer field.}
  \label{fig:partition}
\end{figure}

At a general level the system groups opponent teams by similarity in the player 
formation. In order to understand the player formation, the distribution of the 
players is used. In this paper, offensive corner-kick formations were designed 
regarding the defensive formation of the opponent. Therefore, 
this work suggests to build such player distribution representing the defense of
the opponent by considering locations of players over the corner-kick area. As a
way to represent the opponent player distributions, the system designs a 
partition of the corner-kick area of the field as shown in \figurename 
\ref{fig:partition}. This partition is totally arbitrary, but shows how opponent 
players are spread in this area during their defensive corner-kick situations. 
Resulting distributions represent the number of players in each of the 18 blocks
in the area of interest (the so-called attacking third). Also, an additional 
block representing the remaining part of the field is considered. For example, 
\figurename \ref{fig:partition} shows eleven opponents in their defensive 
corner-kick formation. By analyzing this defense player formation, the resulting
distribution would be written as the following 19-dimensional integer vector:
\\$[1, 0, 1, 0, 0, 0, 1, 2, 1, 1, 0, 0, 1, 0, 0, 1, 0, 0, 2]$.

If we consider a rougher partition of the field, the player distributions would 
tend to be the same regardless the opponent team. For instance, let us consider 
the extreme case where the grid is only constitued of one cell. By doing so, any 
team would be represented by the 1-dimensional integer vector. Conversely, a 
finer partition would make the player distributions become much more different 
from each other.

\subsection{Clustering process}

Once all opponents' distributions are determined, the degree of similarity 
between each possible pair is analyzed in order to generate a distance matrix. 
The distances between distributions are computed by using the Earth Mover's 
Distance (EMD) \cite{emd} method. EMD provides a pseudo metric measure between 
two probability distributions. It can handle vectors with different 
dimensionalities and weighted features. The measurement process is expressed as 
a transportation problem where one distribution is the supplier and the other 
the customer. The cost between the supplier and customer is related to the 
distance between features of the two distributions which are computed by using a
ground distance such as the euclidean distance. This is an advantage of
using EMD since we can evaluate how much two formations of players are 
different by using a ground distance that makes sense in the case of soccer 
field. Also, the possibility to consider weighted features could become an 
advantage in future work since it is possible to give more importance to certain
parts of the formations.

It is possible to apply hierarchical clustering on the resulting distance matrix
in order to determine clusters of similar opponent teams. This process merges 
the pairs with the smallest distance together until all the opponents belong to 
a single cluster. By using a threshold representing the maximum distance 
accepted between clusters before merging, the user can stop the clustering 
process and then obtain several clusters rather than a single one.

\section{Strategy Selection}
\label{sec:learning}
\subsection{Performance evaluation of player formations}

\begin{figure}[h]
  \centering
  \includegraphics[scale=0.25]{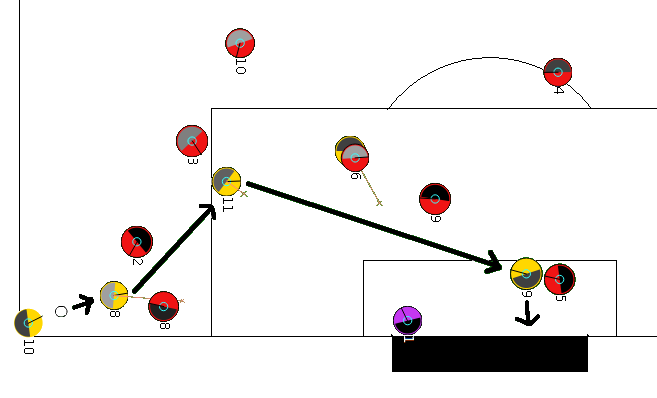}
  \caption{Example of an actions' chain for a corner-kick which leads to a 
successful score.}
  \label{fig:successck}
\end{figure}

In order to select the most effective strategy from a given set of strategies, 
the performance evaluation of the player formations with respect to a success 
metric is required. For example, the probability of success of an attack 
following a corner-kick as shown in \figurename \ref{fig:successck}, can be used
as a performance metric. However, the RoboCup 2D soccer simulation league 
introduces randomness in the way the players interact with the environment. Each
player receives imperfect and noisy information from his virtual sensors. As a 
result, two soccer games with the exactly same teams can differ significantly. 
Therefore, evaluating player positioning performance is a challenging task. 
There is a lot of variance when trying to estimate a success metric. Thus, it is
necessary to run a large number of soccer games in order to estimate one player 
formation's performance with enough precision.

In order to sort each player formation with respect to the others, the 
difference in means between the probability of successful corner-kick 
distributions of each player formation's simulation is considered.

\subsection{Sequential Bayes' estimation}

Bayes' theorem is stated as in (\ref{eq:bayestheorem}):

\begin{equation}
\label{eq:bayestheorem}
 p(\theta | D) = \frac{p(D | \theta)P(\theta)}{p(D)},
\end{equation}

\noindent where $p(\theta | D)$ is called the posterior, $p(D | \theta)$ is 
likelihood, $p(\theta)$ the prior and $p(D)$ is the evidence which stands as a 
normalizing constant. It is calculated as expressed in (\ref{eq:evidence}):

\begin{equation}
\label{eq:evidence}
 p(D) = \int p(D | \theta)p(\theta)d\theta,
\end{equation}

\noindent where $\theta$ represents the value of the parameter to estimate, 
in our case that is the probability of the success of an attack following a 
corner-kick. $D$ corresponds to the new data extracted at the moment 
of applying the theorem. The purpose of the Bayes' theorem is to 
update the prior belief $p(\theta)$ we have about the value of $\theta$ using 
new data $D$. The posterior distribution $p(\theta | D)$ will then correspond to 
our updated belief in the different possible values of $\theta$.

It is possible to sequentially update the parameters by applying Bayes' 
theorem each time one or more simulations are over by using the previous 
posterior as the prior for the next computation of the posteriors.

Obviously, according to the success metric used by the system, the results of 
one experience (successful corner-kicks observed within one game), the 
likelihood follows a binomial law as in (\ref{eq:binomlaw}):

\begin{equation}
\label{eq:binomlaw}
 p(X = k) = C_{k}^{n}\theta^{k}(1 - \theta)^{n - k},
\end{equation}

\noindent where $n$ is the number of total corner-kicks observed during the 
simulated game, $k$ the number of successful corner-kicks observed and $\theta$ 
the probability of an offensive corner-kick to be successful by using the 
player formation. 

Navarro and Perfors \cite{beta-binomial} have demonstrated that the posterior 
distribution of a beta-binomial distribution is also a beta-distribution. Thus,
if you consider the probability of getting a successful corner-kick by using a 
particular formation of players, the posterior distribution after 
observing $k$ successes over the total $n$ corner-kicks can be expressed as in 
(\ref{eq:posterior}):

\begin{equation}
 \label{eq:posterior}
 p(\theta | k, n) \sim B(a + k, n - k + b)
\end{equation}

\noindent where $B$ denotes the beta distribution, $a$ and $b$ are the 
parameters coming from the prior distribution and $\theta$ is the probability of
a successful attack following a corner-kick which is the parameter we want to 
estimate. This fact simplifies computations since it is possible to represent 
the performance of a player formation by a Beta distribution and then after 
running a game, construct a new Beta distribution by giving the number of 
corner-kicks and the number of observed successes.

\subsection{Player formations comparisons}
\label{sec:comparisons}

A difference distribution is used to determine whether one player formation is 
better than another or whether additional simulations are required to be sure. 
For this purpose, the system begins by computing the Highest Density Interval
(HDI) \cite{bayesbook} which is the interval that spans most of the mass of the 
distribution (say 95\%) such that every point inside the interval has a higher 
probability than any point outside the interval.

To compare the performance of two player distributions in the attack case (let 
us say Distribution 1 and Distribution 2), the probability of success of 
Distribution 1 and Distribution 2 is considered, defined as $p_{1}$ and $p_{2}$,
respectively. Assume that a posterior distribution for each of those 
probabilities is obtained. In this case, by calculating all of the possible 
values of $p_{1} - p_{2}$, it is possible to obtain a distribution of the 
difference of $p_{1} - p_{2}$. HDI is used instead of the posteriors in order to
simplify the computation of this calculation.

Then, there are three possible scenarios as follows. Preliminary, let us define 
$[u, v] = \{x \in \mathbb{R} | u \leq x \leq v\}$ to be the HDI of the resulting
distribution $p_{1} - p_{2}$. The first possible case is when $u \geq 0$, which 
means $p_{1} - p_{2} > 0 \Rightarrow p_{1} > p_{2}$. Naturally, the opposite 
case is also possible, $p_{1} - p_{2} < 0 \Rightarrow p_{1} < p_{2}$, which 
happens when $v \leq 0$. Another possible sketch is that 
$[u, v] = \{x \in \mathbb{R} | w \leq u \leq x \leq v \leq z\}$ where $w$ and 
$z$ are around 0 which is equivalent to saying that $p_{1} = p_{2}$ for all 
practical purposes. The $[w, z] = [-0.015, +0.015]$ interval is used in this 
paper. If the two player formations are deemed equal or when the maximum number 
of simulations is reached, the player formation with less variance is considered
as better than the other.

\section{Experiments}
\subsection{Opponents clustering}

First experiments involved 12 teams participating in Japan Open competitions, as 
well as two versions of Agent2D \cite{heliosbasepkg} which does not participate 
in any competitions, but are used by most of the  participants as the starting 
point of team development. Three clusters were created by the hierarchical 
clustering. The second cluster is the most populated among the three ones 
because it represents the teams using a player formation similar 
(if not the same) to that of Agent2D, which constitutes probably their 
implementation starting point. On the other hand, the third cluster included 
only the team Ri-one\_B 2015 that is too far to be merged with any other 
clusters.

\subsection{Association learning}
\label{sec:associationlearning}

In order to experiment the abilities of the learner, we used three corner-kick 
formations that were already implemented in our team. Additionally, a
special script was used. This script runs simulations which only perform 
corner-kick situations. Generally, 37 corner-kicks are executed during one 
simulation, but this number can vary from one run to another due to the 
randomness present in simulations. As first experiment, 10 simulations per 
strategy were simulated before comparing pairs of player formations and a beta 
distribution with parameters 2 and 2 (i.e., Beta(2, 2)) was used to represent 
our prior beliefs.

\begin{figure}[b]
    \centering
    \subfloat[Cluster 1]{{\label{fig:postc1g10}\includegraphics[scale=0.21]{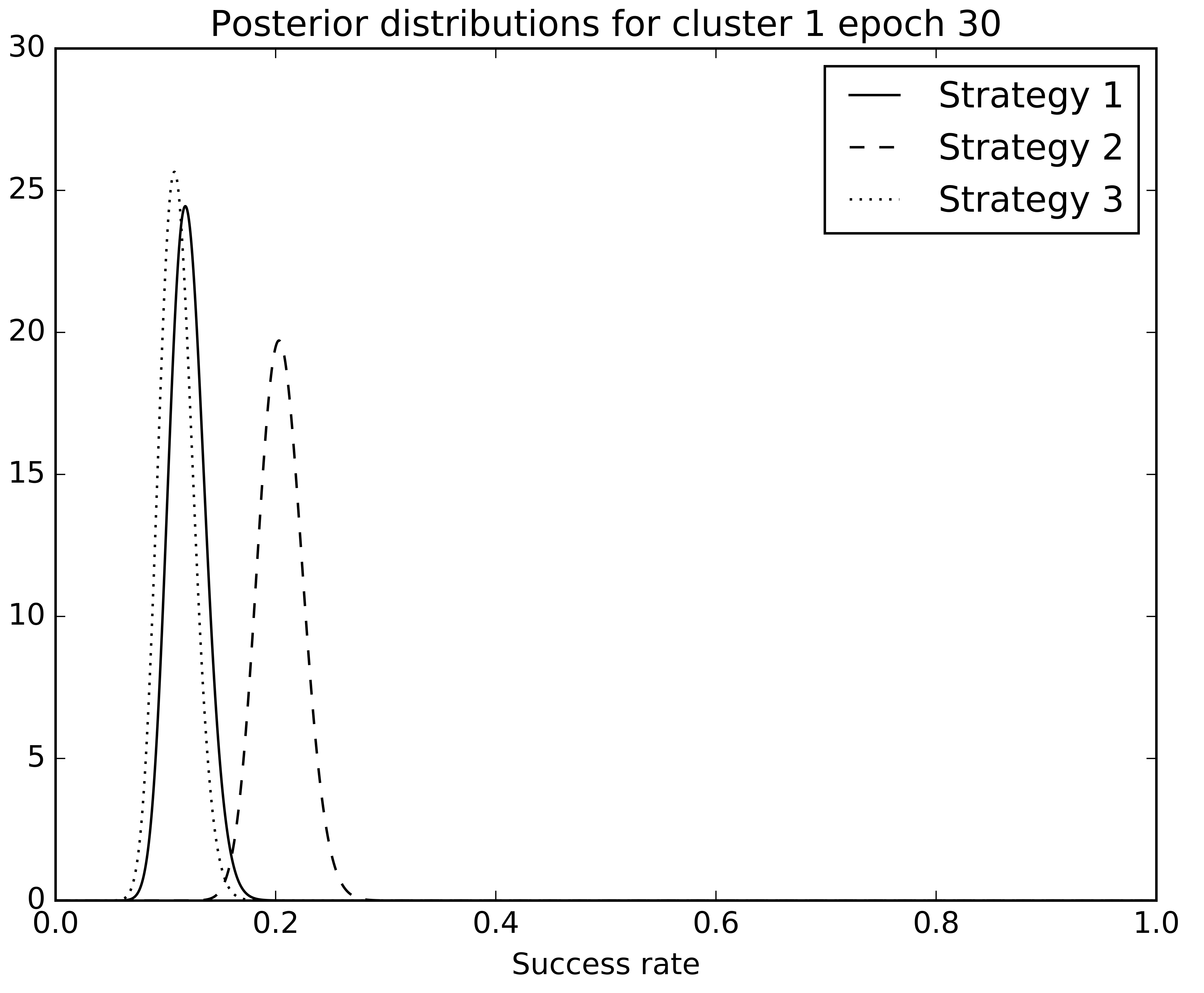}}}
    \qquad
    \subfloat[Cluster 2]{{\label{fig:postc2g10}\includegraphics[scale=0.21]{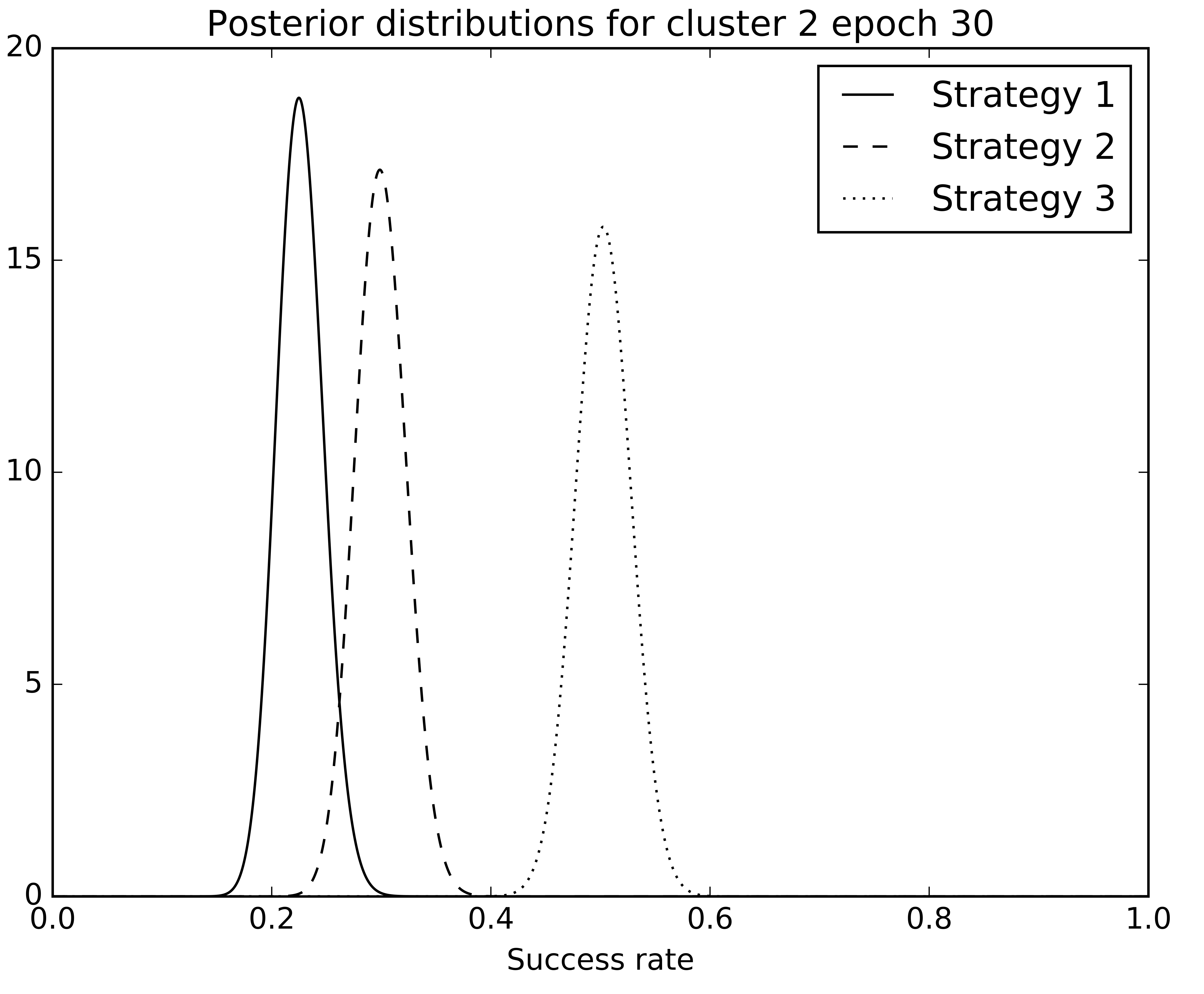}}}
    \qquad
    \subfloat[Cluster 3]{{\label{fig:postc3g10}\includegraphics[scale=0.21]{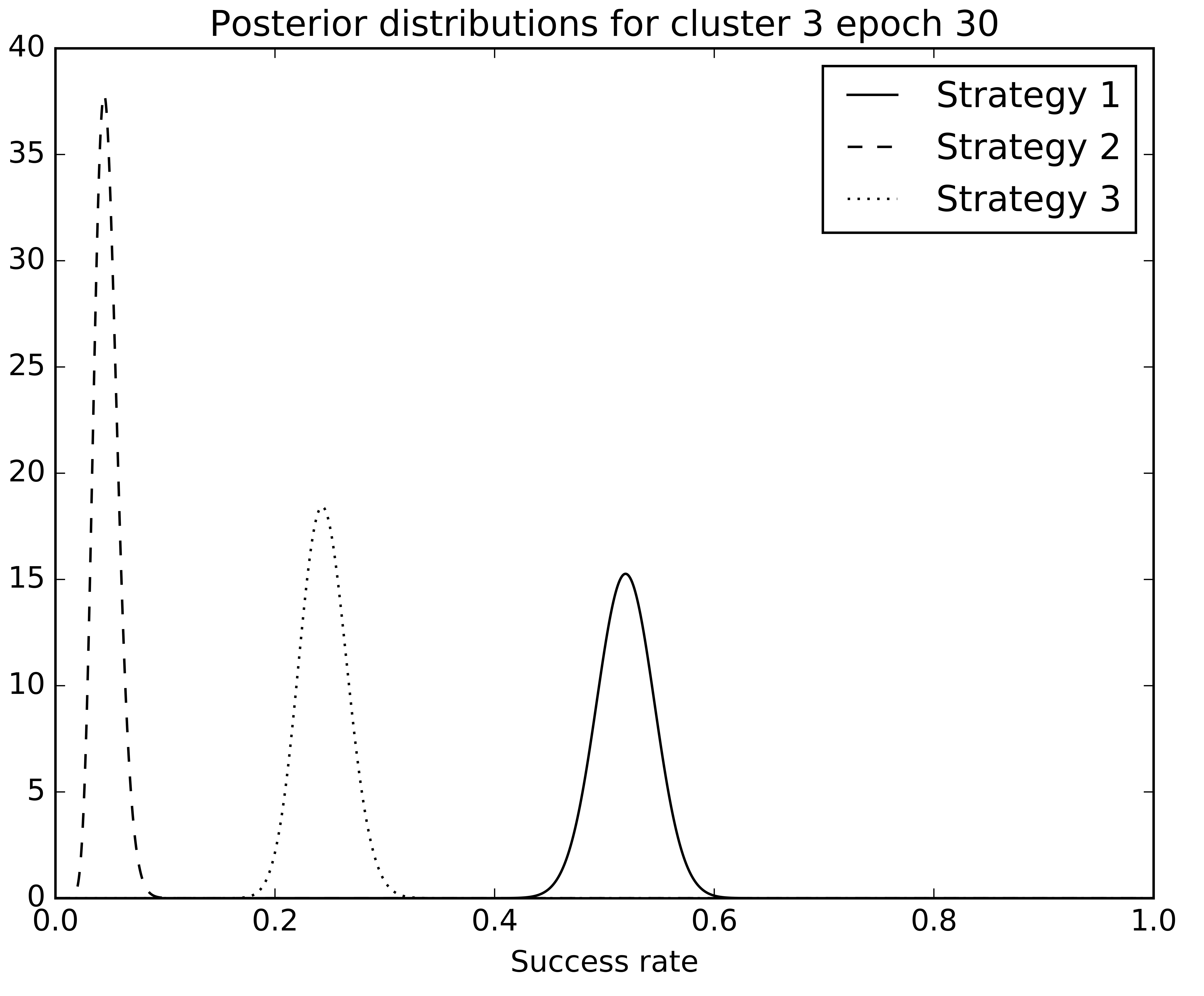}}}
    \caption{Posterior distributions for each cluster, by running $M=10$ simulations}
    \label{fig:postg10}
\end{figure}

The results of our first experiment are shown in \figurename \ref{fig:postg10}, 
the probability density functions of each player formation against each 
cluster. It can be seen that each cluster is associated with a different player 
formation. Excepted the pair (1, 3) for the first cluster (\figurename 
\ref{fig:postc1g10}), all pairs  can be easily ranked. Then, the most effective 
player formation can be determined with certainty. However, the HDI of 
almost all distributions is quite large, thus a precise probability of success
cannot be provided.

\begin{figure}[b]
    \centering
    \subfloat[Cluster 1]{{\label{fig:postc1g60}\includegraphics[scale=0.21]{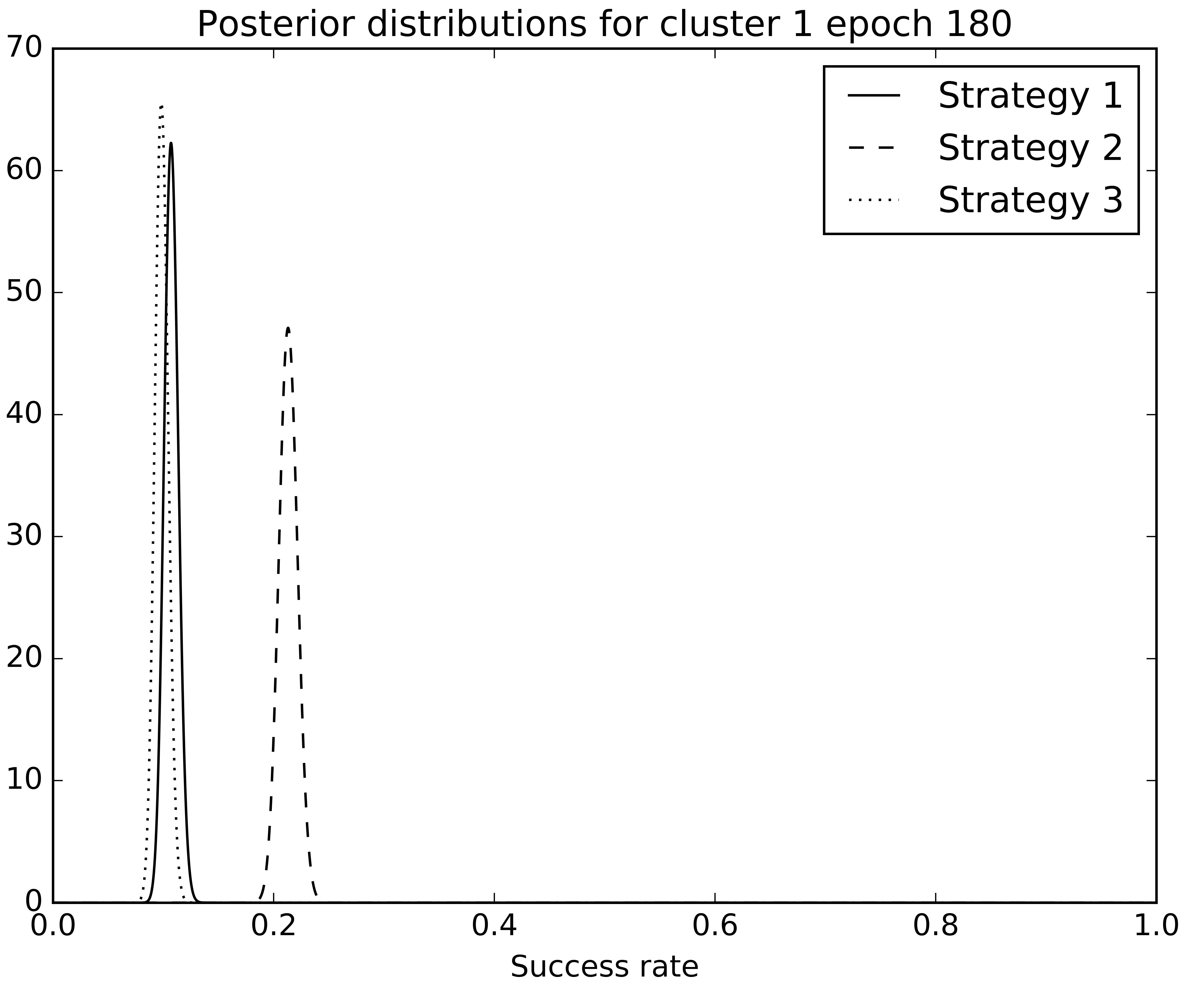}}}
    \qquad
    \subfloat[Cluster 2]{{\label{fig:postc2g60}\includegraphics[scale=0.21]{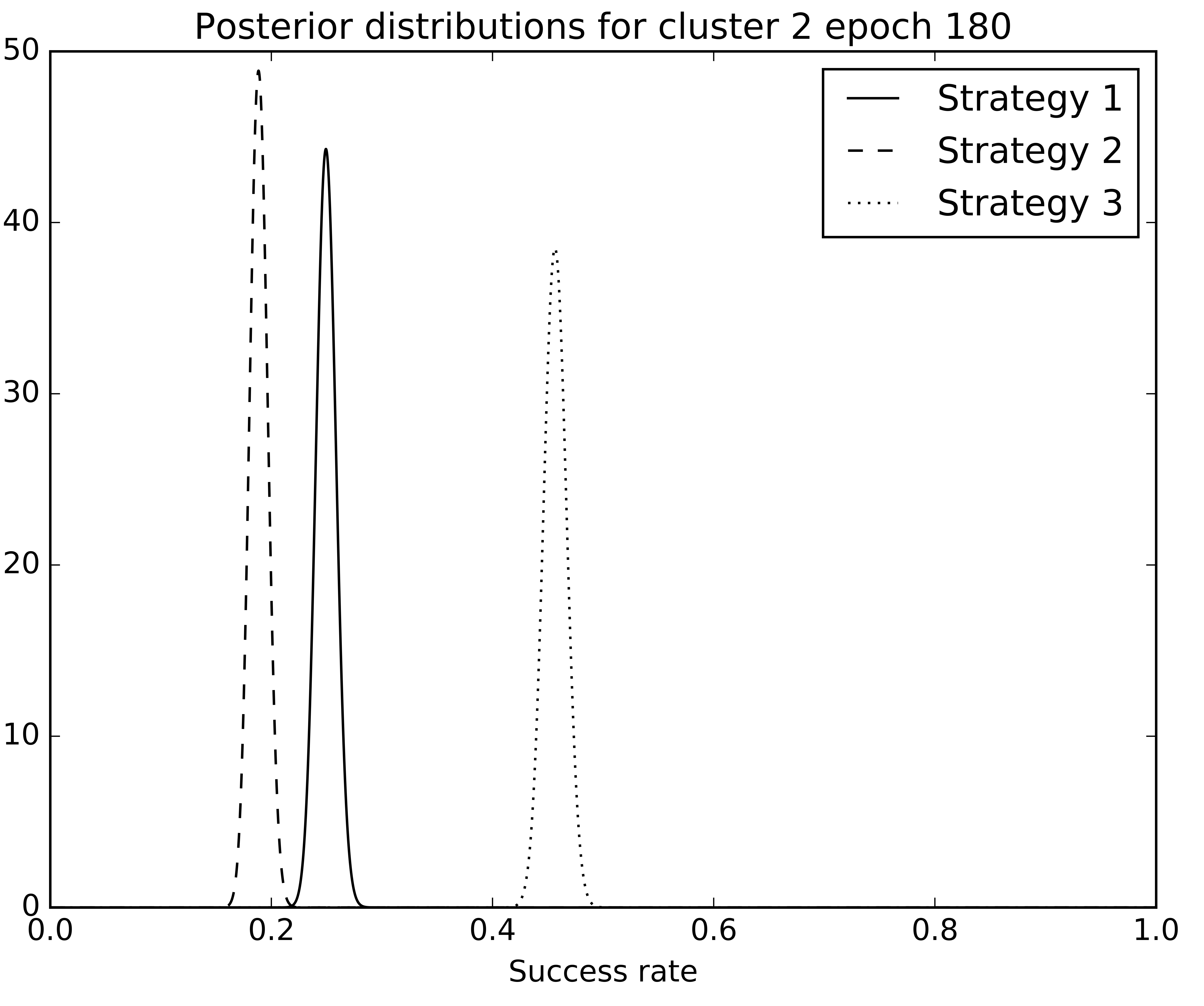}}}
    \qquad
    \subfloat[Cluster 3]{{\label{fig:postc3g60}\includegraphics[scale=0.21]{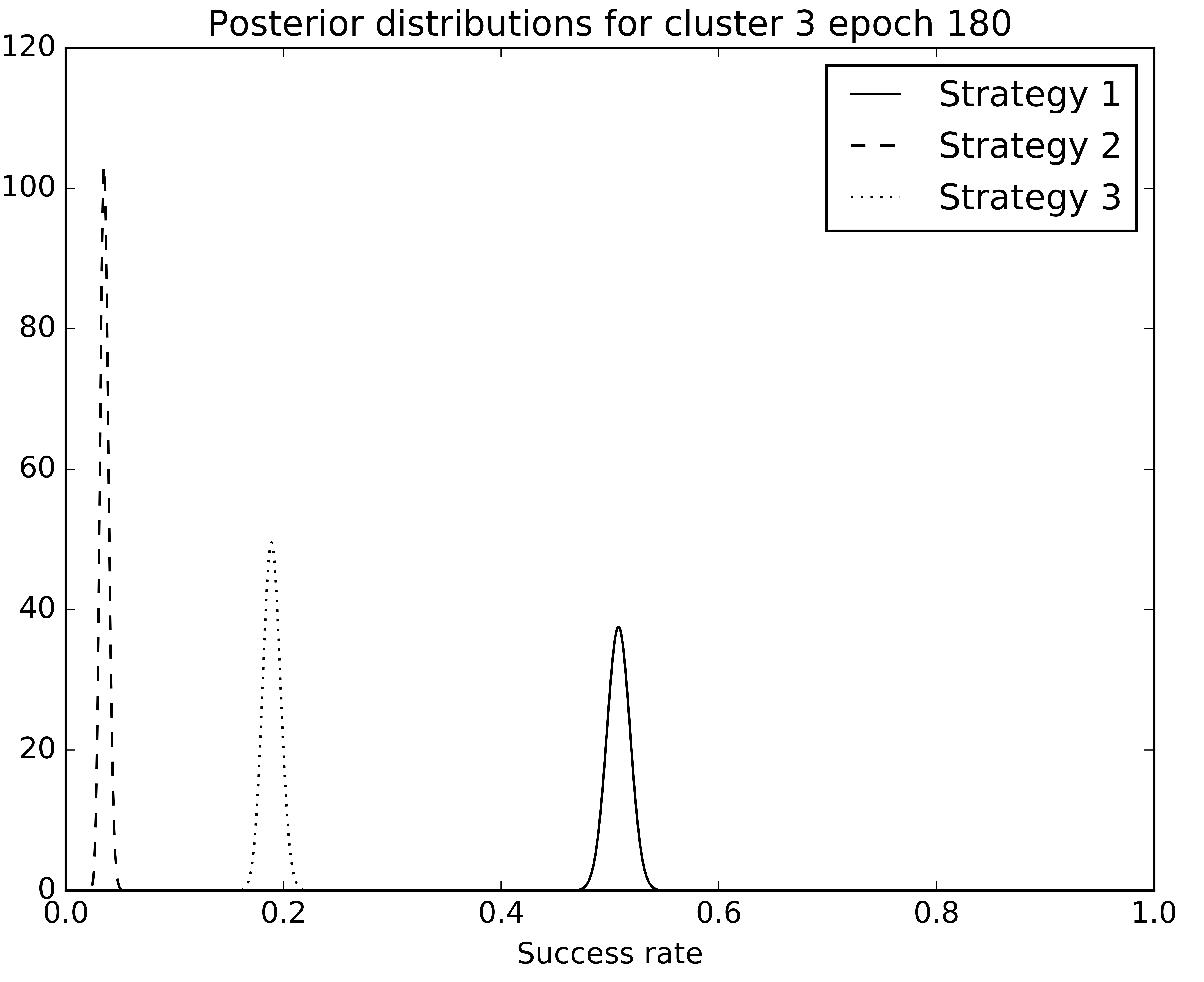}}}
    \caption{Posterior distributions for each cluster, by running $M=60$ simulations}
    \label{fig:postg60}
\end{figure}

In order to improve estimations about the player formations' probability 
of success, a second experiment was conducted and simulations were generated by 
blocks of 60 games. That is, 60 games for each player formation were 
conducted every time the performance is compared. \figurename \ref{fig:postg60} 
shows the resulting probability density functions of the player formation. 
As expected, curves became finer and tended to be centered to the true 
probability of their respective player formation. Also, the pairs which were 
difficult to differentiate after the first experiment, can now be well ordered.

Table \ref{tab:resultssummary} provides a summary of the second experience. It 
shows the final associated player formation for each cluster. Also, it 
indicates the HDI of the selected player formation. Finally, it gives the ratio
of the best player formation's distribution mean over the second best's.

\begin{table}[!t]
\caption{Results summary of the second experiment.}
\label{tab:resultssummary}
\centering
 \begin{tabular}{|c|c|c|c|}
 \hline
 Cluster & Distribution & HDI & Ratio\\
 \hline
 1 & 2 & [0.203, 0.237] & 1.787\\
 \hline
 2 & 3 & [0.531, 0.571] & 2.073\\
 \hline
 3 & 1 & [0.471, 0.512] & 2.179\\
 \hline
 \end{tabular}
\end{table}

\subsection{System validation} 

The proposed method in this paper estimates the probability of success of 
offensive player formations against given opponent teams. In other words, the 
parameter $\theta$ of a binomial distribution is estimated. However, it is 
legitimate to wonder about the correctness of the estimations.

The experiment in this section puts player formations aside and evaluates how 
well our method can differentiate probability distributions with parameters 
close together. Additionally, it estimates how many simulations are required to 
draw trustful conclusions about the ranking of offensive player formations 
regarding their success probability.

Player formations are substitued by a set of randomly generated parameters 
$\theta$. Then, a simulation of $n$ offensive corner-kicks by using a particular
formation is substitued by $n$ sampling from a binomial distribution 
parameterized by one of the randomly generated $\theta$ values. Notice that the 
system knows the generated parameters and is able to order them. Afterwards, as 
in the parameter estimation method, by feeding the Bayesian estimator with the 
number $k$ of successes over the $n$ samples the system updates prior beliefs 
about the parameters and tries to estimates the value of the randomly generated 
parameters. Actually, since the true values are known, it is possible to verify 
that the system gets back the correct ranking.

\begin{figure}[!t]
  \centering
  \includegraphics[scale=0.34]{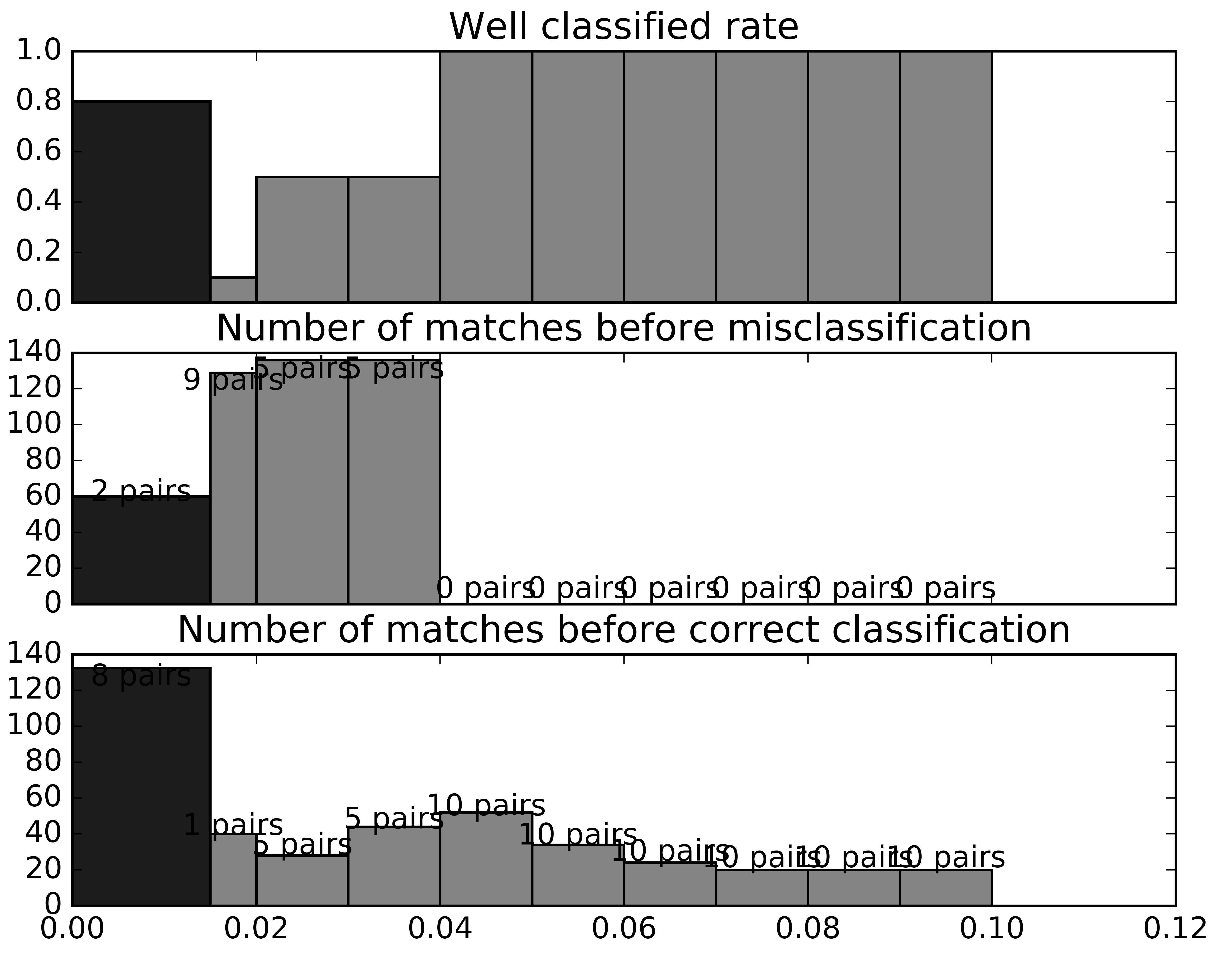}
  \caption{System's validation by blocks of 20 samples.}
  \label{fig:validation20}
\end{figure}

\figurename \ref{fig:validation20} shows the results obtained by using $n = 20$
samples per simulation for each distribution. The figure consists of three 
subplots where the $x$-axis represents bins of pairs of $\theta$s depending on 
the difference of their respective values. For example, assume 
$\theta_{1} = 0.22$ (22\% of chance to get a success) and $\theta_{2} = 0.24$. 
Since the difference between $\theta_{1}$ and $\theta_{2}$ is 0.02, this pair is
contained in the second bin whose range is from 0.1 to 0.2. The first bin (the 
one in black) is special, since it represents the interval where parameters are 
close enough to be considered equal. The number of generated parameters was done
in such a way that each bin contains ten pairs of parameters.

The first subplot shows the rate of well ordered pairs in each bin. According
to this plot, the system can perfectly rank pairs with a difference greater than
0.04 and this accuracy decreases as the distances between parameters increase. 
In this subplot the correct ranking inside the first bin is not really important
since it contains pairs that are considered to be equal.

The second and third subplots show the number of correctly ranked (respectively
uncorreclty ranked) pairs in each bin and the number of sampling steps before 
drawing conclusion ($y$-axis). As indicated in the first subplot, the ranking is
perfect for any pairs contained in the bin whose range is greater than $0.04$. 
Furthermore, at most fifty samples were required to obtain such results and this
number decreases as the distances increase. However, the system has difficulties
to rank pairs with difference less than $0.04$. 

\begin{figure}[!t]
  \centering
  \includegraphics[scale=0.34]{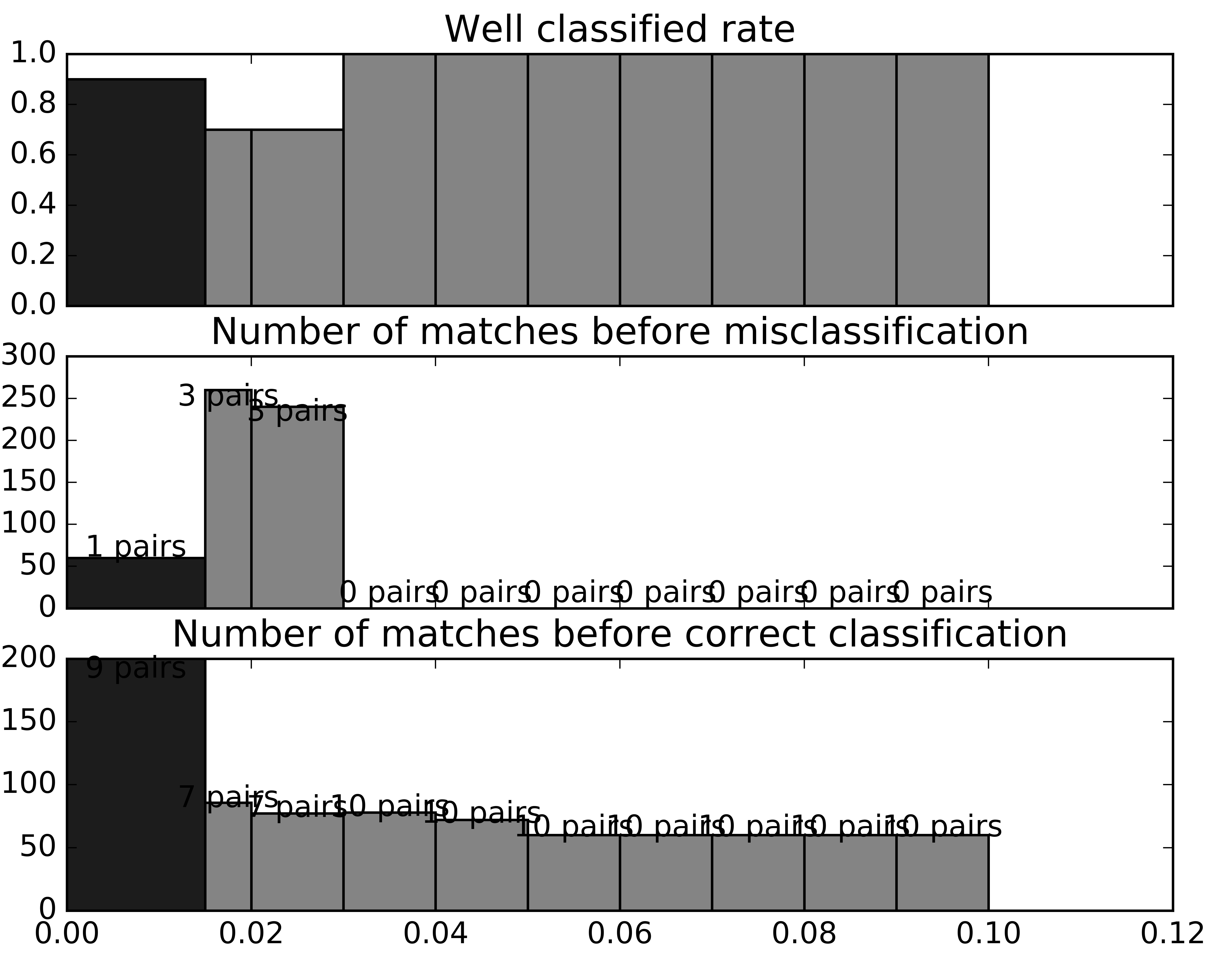}
  \caption{System's validation by blocks of 60 samples.}
  \label{fig:validation60}
\end{figure}

\figurename \ref{fig:validation60} shows the performance evaluation by using 
$n = 60$ samples per simulation for each distribution. Actually, increasing this
number improves the accuracy since the system is able to rank prefectly pairs 
with at least a difference of $0.03$ by requiring at most eighty samples. These 
two experiments show that increasing the number of samples increase the 
accuracy. On the other hand, since the data that the player receives is biased
and because of the rarity of corner-kick event occurence during one single game,
a deviation of 4\% of success probability between two formations is not so 
significant. For this reason, in the particular case of selecting the best 
strategy for offensive corner-kicks, estimating the formations' parameter by 
running only 20 simulations is enough.

\subsection{Cluster validation}

It could be also interesting to look at the performance of each player 
formation in the clusters. While some teams have been considered as similar in 
terms of defensive player formations, it does not exclude the possibility of 
disparities among the teams of the same cluster since results of actions are 
not affected by player positioning only. Proper agents' skills are an equally 
important factor.

In order to verify the quality of association according to each opponent team, 
another alternative of the algorithm was applied. This one is nearly the same as 
the standard version, but rather than trying to estimate the effectiveness of 
each player formation against the clusters, the system estimates it against 
every team individually.

Table \ref{tab:differenceexpectations} summarizes the teams whose the most 
effective strategy is not the same as the one estimated against the cluster 
which they belong. As a reminder, Cluster 1 counts three teams and Cluster 2 
counts ten teams. Regarding the team Ri-one\_A 2015 (Cluster 1), Distribution 1 
seems to be better than Distribution 2 which is the one associated to its 
cluster. However, the error seems to be much more serious, according to the team
A\_TSU\_BI-2014 (Cluster 2) since Distribution 2's mean is slightly more than 
three times better than the selected formation's (Distribution 3) mean. In fact, 
this association error is not significant during a game against Ri-one\_A 2015. 
On the other hand, performing games against A\_TSU\_BI-2014 with the wrong 
strategy would affect the results of games since there is roughly 20\% more 
chance to get a success by using the formation associated to Distribution 2
rather than the one selected (Distribution 3).

\begin{table}[!h]
\caption{Difference between expected performances in cluster.}
\label{tab:differenceexpectations}
\centering
 \begin{tabular}{|c|c|c|c|c|}
 \hline
 Team & Cluster & Selected (Dist. / HDI) & Best option (Dist. / HDI) & Ratio\\
 \hline
 Ri-one\_A 2015 & 1 & 2 / [0.13, 0.16] & 1 / [0.23, 0.27] & 1.73\\
 \hline
 A\_TSU\_BI- 2014 & 2 & 3 / [0.07, 0.09] & 2 / [0.23, 0.26] & 3.28\\
 \hline
 \end{tabular}
\end{table}

\section{Conclusion}

In this research, a system that is able to select the best player formation 
in corner-kick situations regarding a group of teams was developed. This 
decision is taken by doing sequential Bayes' estimations from the results of 
several games. The model does not create effective offensive player formations, 
but instead indicates the best that we have already in hand.

The results are satisfying since the system is able to rank correctly 
player formations with at least a difference of 4\% of success probability 
by proceeding only 20 simulations. Furthermore, it is possible to increase the 
precision of the system by getting more data. However, by doing so the learning 
time would increase considerably. Additionally, it is quite impossible to feel a 
difference during one game, since during a true match the number of 
corner-kicks that happen is very low. This is why such an error rate is 
acceptable.

On the other hand, there is a possibility of disparities inside the clusters. As
explained earlier, if the difference between player formations is only 4\% 
there is actually no real difference in terms of final results of one game due 
to the rare occurence of corner-kicks executed during a standard game. But if a 
player formation is not designed to be the best and is actually three times 
better than the selected one, difference could be observed regarding final 
results. These disparities are due to the fact that during the clustering 
process only the positions of opponents are considered and not the defensive 
skills of the team. Then, another clustering criterion can be considered for 
better performance.

Finally, while the first trials selected player formations for corner-kicks 
only it is possible to use it for any situation of the game, at the condition 
to have a criterion for opponents clustering and a success metric for data 
observations. Furthermore, it is possible to extend this system in order to 
build strategies, i.e. sets of player formations that cover any situation, 
rather than selecting the best player formation according to a particular 
situation. In this case opponents would not be in only one cluster, but in 
several clusters, one for each situation. However, such a learning process 
seems difficult to realize since a very large number of standard games is
required, ones which do not simulate only one kind of situation, in 
order to see enough every kind of situations and hope to obtain good 
approximations of each player formation.

\end{document}